\documentclass[runningheads]{llncs}
\usepackage[T1]{fontenc}
\usepackage{graphicx,verbatim}
\usepackage{algorithm}
\usepackage{algpseudocode}
\usepackage{url}
\usepackage{graphicx}
\usepackage[table,xcdraw]{xcolor}
\usepackage{verbatim}
\usepackage{hyperref}
\usepackage{booktabs}
\usepackage{cprotect}
\usepackage{xspace}
\usepackage{tikz}
\usepackage{arydshln}
\usepackage[toc,page,titletoc]{appendix}
\usepackage{rotating}
\newcommand{\tcga}{\textbf{Midnight-12k}\xspace}
\newcommand{\nki}{\textbf{Midnight-92k}\xspace}
\newcommand{\pt}{\textbf{Midnight-92k/392}\xspace}
\algblock{ParFor}{EndParFor}
\algnewcommand\algorithmicparfor{\textbf{parfor}}
\algnewcommand\algorithmicpardo{\textbf{do}}
\algnewcommand\algorithmicendparfor{\textbf{end\ parfor}}
\algrenewtext{ParFor}[1]{\algorithmicparfor\ #1\ \algorithmicpardo}
\algrenewtext{EndParFor}{\algorithmicendparfor}
\newcommand{\size}[1]{#1$\times$#1}
\begin{document}
\title{Training state-of-the-art pathology foundation models with orders of magnitude less data}
\titlerunning{Training state-of-the-art pathology FMs with orders of magnitude less data}
\author{Mikhail Karasikov \inst{1} \and
Joost van Doorn \inst{1} \and
Nicolas Känzig \inst{1} \and
Melis Erdal Cesur \inst{2} \and
Hugo Mark Horlings \inst{2} \and
Robert Berke \inst{1} \and
Fei Tang \inst{1} \and
Sebastian Otálora \inst{1}}
\authorrunning{Karasikov et al.}
\institute{Kaiko.AI, Zurich, Switzerland \and
The Netherlands Cancer Institute, Amsterdam, The Netherlands\\
\email{\{mikhail,sebastian\}@kaiko.ai}}
\maketitle
\begin{abstract}
The field of computational pathology has recently seen rapid advances driven by the development of modern vision foundation models (FMs), typically trained on vast collections of pathology images.
Recent studies demonstrate that increasing the training data set and model size and integrating domain-specific image processing techniques can significantly enhance the model's performance on downstream tasks. Building on these insights, our work incorporates several recent modifications to the standard DINOv2 framework from the literature to optimize the training of pathology FMs. We also apply a post-training procedure for fine-tuning models on higher-resolution images to further enrich the information encoded in the embeddings.
We present three novel pathology FMs trained on up to two orders of magnitude fewer WSIs than those used to train other state-of-the-art FMs while demonstrating a comparable or superior performance on downstream tasks.
Even the model trained on TCGA alone (12k WSIs) outperforms most existing FMs and, on average, matches Virchow2, the second-best FM published to date. This suggests that there still remains a significant potential for further improving the models and algorithms used to train pathology FMs to take full advantage of the vast data collections.
\keywords{Foundation models  \and Computational pathology \and Whole Slide Images.}
\end{abstract}
\section{Introduction}
Recently, there has been an increased interest in developing vision foundation models for various types of images, including medical imaging data.
These FMs generate informative representations that can be used in various downstream tasks such as classification, segmentation, object detection, etc.
In particular, the development of foundation models for computational histopathology, commonly referred to in the literature as {\em pathology FMs}, has rapidly accelerated~\cite{OKI2025}.

This progress has been primarily driven by the ever-increasing amount of unlabeled Whole Slide Image (WSI) data available in public and proprietary sources, the development of more complex model architectures (e.g., ViT~\cite{dosovitskiy2020image}), and the steady refinement of training workflows (e.g., DINOv2~\cite{ODM2023}).
Most current state-of-the-art pathology FMs are based on either the DINO[9] or DINOv22~\cite{ODM2023}) self-supervised learning (SSL) algorithms. DINOv2 was developed as an extension of DINO~\cite{caron2021emerging} and iBOT\cite{zhou2021ibot} to train general-purpose vision FMs that capture both the global context and local structure of the images, using Vision Transformers (ViTs)~\cite{dosovitskiy2020image}  as an underlying image encoder.

Depending on the type of input images, pathology FMs can be designed to produce tile- or slide-level representations.
In this work, we only consider tile-level FMs.
For training a tile-level FM, the original WSIs are pre-processed to extract smaller regions, {\it tiles} (also often referred to in the literature as {\it patches}), typically of size \size{224} pixels.

One of the recent milestones in the development of pathology FMs was UNI~\cite{chen2024uni}, a ViT-L16 model trained with DINOv2 on over 100k WSIs from various sources, where the authors set a new standard in the performance of a pathology FM and conducted numerous experiments evaluating it on diverse downstream tasks.
Very recently, the same group released a successor model UNI-2~\cite{uni2_huggingface} trained on over 200M tiles sampled from over 350k diverse H\&E and IHC WSIs.
In Kaiko-FM~\cite{KAJ2024}, the authors trained relatively performant models solely on TCGA and introduced {\it online patching}, an efficient technique for sampling WSI tiles of arbitrary size directly during training to reduce the space overhead.
H-optimus-0~\cite{SJL2024} is a ViT-g14 trained with DINOv2 on 500k WSIs with several hundreds of millions of tiles. Their model is one of the largest in terms of the number of parameters and still remains one of the best-performing published models according to various benchmarks.
Another prominent model is Virchow2~\cite{ZVV2024}, which is a ViT-H14 trained on a substantially larger data set comprising 3.1M WSIs.
In Hibou~\cite{NPI2024}, the authors trained a family of FMs with 936,441 H\&E, 202,464 non-H\&E, and 2,676 cytology slides sourced from 306,400 unique cases.
Very recently, Atlas~\cite{ATD2025} was released as a preprint, where the authors trained a new pathology FM and demonstrated outstanding performance on the HEST~\cite{JDS2024} benchmark and six out of eight downstream tasks from \textit{eva}~\cite{GKM2024}. However, without released weights, external evaluation of that model appears impossible.

Training pathology FMs at large scale~\cite{ZVV2024,XUB2024,SJL2024,NPI2024,chen2024uni,KAJ2024,NPI2024,ATD2025} have pushed the frontier by amassing tens of thousands to millions of WSIs from both public and proprietary sources.
However, the tendency towards scaling leaves a critical open question: is it crucial to have such a large data set in order to train a pathology FM at the state-of-the-art level, or can similar results be achieved with far fewer WSIs?

In this work, we address the question posed above and present three novel pathology FMs trained on relatively small publicly available data sets and a proprietary set of over 80k WSIs from the Netherlands Cancer Institute (NKI). Despite being trained on orders of magnitude fewer WSIs than most other state-of-the-art models published to date, our models achieve comparable and often higher performance on most downstream tasks.
We additionally perform an ablation study to determine the contributions of the individual changes we made to the standard DINOv2 training workflow.
Drawing an analogy between the whole range of models applied to data at different scales (from single molecules, to cells, to tissue samples, and to entire organisms) and depth zones of the ocean, we call our pathology FMs after the middle depth zone in the ocean, bathypelagic ({\em midnight}) zone.
The shared models and the source code and data necessary to reproduce the evaluation experiments are available at \url{https://github.com/kaiko-ai/midnight}.

\section{Methods}
\subsubsection{Training data}
We trained our FMs and performed the ablation study on three public collections of WSIs: TCGA, GTEx, and CPTAC, and a proprietary data set NKI-80k.
TCGA contains 12k FFPE slides from 32 cancer types collected in different hospitals by the TCGA Research Network: \url{https://www.cancer.gov/tcga}.
GTEx contains 25k WSIs across 23 tissue types from 838 donor individuals, collected by the Genotype-Tissue Expression project~\cite{doi:10.1126/science.1262110}.
CPTAC contains 7.2k WSIs from clinical tumor samples from 13 cohorts collected by the Clinical Proteomic Tumor Analysis Consortium~\cite{NOT2015}.

In addition to those three open-access data sets, we use a proprietary set of 80k WSIs (NKI-80k) from the Netherlands Cancer Institute. These slides have magnifications of 0.25 and 0.5\,µm/px, most of them being at 0.25\,µm/px.
This data set includes mostly FFPE H\&E but also Frozen Tissue and immunohistochemistry slides from 10,141 patients and 31 organs.
In our experiments, we found that including GTEx and CPTAC slides in training did not bring substantial improvements (see Results). Thus, we trained our final FMs only on the TCGA FFPE and NKI-80k slides.
\subsubsection{Extraction of training tiles}
We trained our FMs on tiles of size \size{256} cropped from the original WSIs at magnifications of 2, 1, 0.5, and 0.25\,µm/px.
All tiles were sampled uniformly at random from arbitrary positions of the WSIs with {\it online patching}~\cite{KAJ2024}, with the foreground area threshold set to 40\%.
Further, to filter out low-informative tiles (e.g., those with mainly adipose tissue), we apply a filter in the HSV color space from~\cite{ZVV2024}.
More precisely, a tile is only accepted if $\geq$60\% of its pixels have their hue, saturation, and value in ranges [90, 180], [8, 255], and [103, 255], respectively (see examples in Fig.~\ref{fig:img_preprocessing}-Left).
For all cropped tiles, we apply color augmentations in the Hematoxylin-Eosin-DAB (HED) space~\cite{TLB2019} (Fig.~\ref{fig:img_preprocessing}-Right). These augmentations effectively increase the diversity of the training data and help make the FM more robust to various staining methods used in the WSIs.
\subsubsection{Self-supervised training with DINOv2}
We use the DINOv2 self-distillation framework to train ViT-g14 models with 1.1B parameters (and ViT-B14 with 86M parameters in the ablation experiments) with self-supervised learning. Our algorithm is based on the original DINOv2 algorithm~\cite{ODM2023} with several modifications.
First, as suggested in~\cite{ZVV2024}, we use a more stable KDE regularizer~\cite{WIS2020} instead of the original KoLeo loss to ensure the diversity of tile embeddings generated by the FM.
We start from the checkpoints pre-trained in~\cite{ODM2023} and train on 32 Nvidia-H100 GPUs with 80\,GB memory for 1M iterations with the base learning rate of $3.5\times10^{-4}$ and the learning schedules compressed accordingly, with the batch size of 12 per GPU. (In total, it extracts throughout the training $384\times10^6$ tiles from the WSIs.) We accumulate gradients over every two training steps, resulting in an effective total batch size of 768.
\subsubsection{High-resolution post-training}
As in~\cite{ODM2023}, after training, we optionally further fine-tune the FM on larger images for 120k iterations to improve its performance, especially on high-resolution images.
\begin{figure}[t]
    \centering
    \includegraphics[width=1\linewidth]{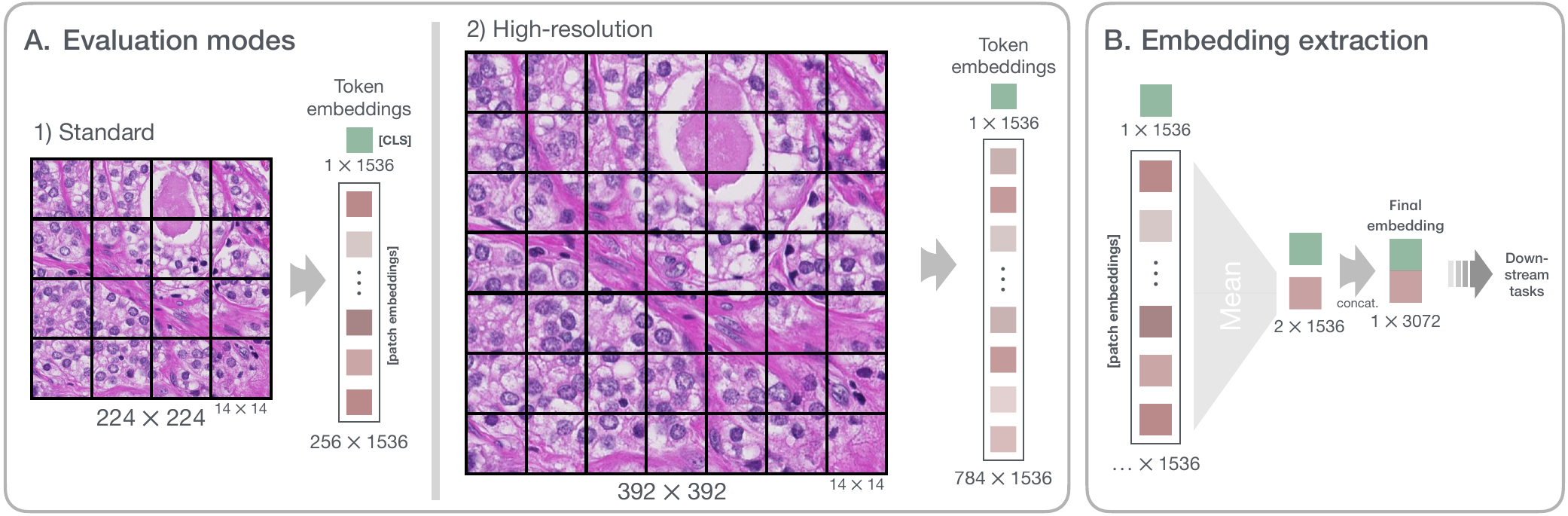}
    \caption{Schematic representation of the FM evaluation. \textbf{Panel A:} Evaluation of a vision transformer FM at standard and high resolution. For high resolution, larger tiles of size \size{392} are cropped into $(392/14)^2=784$ patches of the same size \size{14} pixels. (The grids are shown schematically for simplicity. The actual numbers of patches the tiles are cropped into are $256$ and $784$ instead of $4^2$ and $7^2$, as shown in the graph.)
    \textbf{Panel B:} Aggregating token embeddings produced by the ViT into the final CLS+Mean token embedding.}
    \label{fig:schematic}
\end{figure}
Similar to reducing the patch size in the underlying vision transformer studied in~\cite{beyer2023flexivit}, this technique effectively increases the number of patch tokens generated from every image by the ViT. At the same time, it allows us to start from an FM pre-trained at standard resolution and thereby shorten the training on large images.
More precisely, for this post-training, we increase the size of training tiles from 256 to 512 pixels, and accordingly reduce the magnification by 2-fold, to 1, 0.5, 0.25, and 0.125\,µm/px, to preserve the actual size of the tile regions (512, 256, 128, and 64\,µm).
In addition to increasing the resolution, we also scale up the parameters of the DINO transform from 98 and 224 to 168 and 392 for the local and global crop views, respectively.
Since the use of larger images increases the memory requirements, we reduce the batch size to 6 per GPU and train on 48 GPUs with accumulating gradients over every four training steps, resulting in the effective total batch size of 1152. The base learning rate in this stage is reduced to $10^{-4}$.
Consequently, at inference time, each input image is resized to \size{392} before passing it to the FM for the embedding generation (see Fig.~\ref{fig:schematic}A).
Note that this resizing does not change the actual region of the WSI contained in the tile.
To the best of our knowledge, we are the first to apply this high-resolution fine-tuning procedure for training pathology FMs.
\subsubsection{Training three new FMs}
We trained three new FMs: 1) We trained our first FM on the 12k TCGA WSIs alone, with the training methodology described above.
We refer to this model as \tcga.
2) For our second model, we applied the same training algorithm on both TCGA and NKI-80k combined. (Each batch was sampled from TCGA or NKI-80k at random with equal probabilities.)
We refer to this model as \nki.
3) Finally, we fine-tuned the \nki FM with the high-resolution post-training technique described above, with reduced training schedules, for 120k more iterations. We refer to this model as \pt.
\subsubsection{Evaluation methodology}
Our evaluation protocol is based on two open-source benchmarks: eva~\cite{GKM2024} and HEST~\cite{JDA2024}.
eva includes various tile- and slide-level classification tasks and two tile segmentation tasks, assessing how well the FMs encode tissue morphology in different tissues and cancers.
The original data sets from which the downstream tasks in eva were derived are summarized in Table~\ref{tab:datasets}.
\begin{table}[!b]
\centering
\caption{Data used in the evaluated downstream tasks. All tiles in all tasks are resized to \size{224} (or other respective dimensions) before passing to the FM for computing embeddings. (*) For slide-level tasks Camelyon16 and Panda, the values in columns `Tile size' and `Magnification' represent the tiles cropped from the original WSIs before resizing them to the target dimensions.}\label{tab:datasets}
\scalebox{0.8}{%
\begin{tabular}{|l|rrrrr|}
\hline
\textbf{Task name} & \textbf{\#\,images} & \textbf{Tile size} & \textbf{Magnification} & \textbf{Organ} & \textbf{Metric} \\
\hline
BACH~\cite{PEA2019}         & 400        & 1536x2048 & 0.42\,µm/px (20x)         & Breast     & Bal. acc.   \\
BRACS~\cite{BAP2022}        & 4,539      & variable  & 0.25\,µm/px (40x)         & Breast     & Bal. acc.   \\
BreaKHis~\cite{SOP2015}     & 7,909      & 700x460   & 1.995\,µm/px (40x)        & Breast     & Bal. acc.   \\
CRC-100K~\cite{KHM2018}     & 107,180    & 224x224   & 0.5\,µm/px (20x)          & Colorectal & Bal. acc.   \\
Gleason TMA~\cite{AFM2018}  & 21,496     & 750x750   & 0.23\,µm/px (40x)         & Prostate   & Bal. acc. \\
MHIST~\cite{WSR2021}        & 3,152      & 224x224   & 1.25\,µm/px (8x)          & Colorectal & Bal. acc.   \\
PatchCamelyon~\cite{VLW2018}         & 327,680    & 96x96     & 1\,µm/px (10x)            & Breast     & Bal. acc.   \\
Camelyon16*~\cite{BVV2017} & 399 WSIs        & 224x224   & 0.25\,µm/px (40x)          & Breast     & Bal. acc.   \\
Panda*~\cite{BKC2022}   & 1909 WSIs   & 448x448   & 0.25\,µm/px (40x)          & Prostate   & Bal. acc.   \\
CoNSeP~\cite{GVR2019}        & 41         & 1000x1000 & 0.25\,µm/px (40x)         & Colorectal & Dice score        \\
MoNuSAC~\cite{VKP2021}       & 294        & variable  & 0.25\,µm/px (40x)         & Various    & Dice score        \\
HEST (all)~\cite{JDS2024}         & 236,495    & 224x224   & 0.5\,µm/px (20x)          & Various    & Pearson $\rho$ \\
\hline
\end{tabular}}
\end{table}
For all tasks in eva, the original tiles are resized to the desired dimensions before passing them to the FMs for embedding generation (e.g., cropping the center squares from the original 700$\times$460 tiles at 1.995\,µm/px in BreaKHis and resizing them to \size{224} results in tiles of size \size{224} at 0.97\,µm/px).
For all tasks except Camelyon16 and Panda, we also disabled early-stopping in eva's default protocol to ensure that all evaluation runs fully converge.

The HEST benchmark includes nine tile-level tasks that evaluate how well the FM can predict gene expression from histology images. Each task is a regression of the FM's embeddings of the \size{224} tiles at 0.5\,µm/px to normalized transcript counts of the top 50 highly variable genes, measured at the respective positions of the tiles.
Performance in HEST is measured by the Pearson correlation coefficient between the predicted and actual gene expression, computed across all patients.
\section{Results and Discussion}
\label{sec:results}
\subsubsection{Reaching state-of-the-art performance with less data}
We evaluated the performance of our FMs and several other state-of-the-art FMs on the downstream tasks described above.
For every model, we evaluated both the CLS token and the CLS+Mean token embeddings (the concatenation of the CLS token and the mean of all (image\_size$/$patch\_size)$^2$ patch tokens in the vision transformer, see Fig.~\ref{fig:schematic}B).
For HEST, we only report the aggregate average of Pearson correlations.
It can be seen (Table~\ref{tab:results}) that even our model \tcga trained on just 12k WSIs is superior to most other existing FMs, and is only marginally different from Virchow2 despite being trained on 258$\times$ fewer WSIs (12k vs. 3.1M).
\begin{table}[t]
\caption{Performance metrics for all evaluated FMs on the data sets from Table~\ref{tab:datasets}, and their average.
pc10 is a tile-level classification task derived from PatchCamelyon (pc) where the training set is reduced to just ten random tiles per class (20 tiles in total).
We report balanced accuracy for the classification tasks, dice score (no background) for semantic segmentation (cnsp, mnsc), and the average Pearson correlation for the nine HEST regression tasks. All classification tasks use CLS+Mean embeddings (the concatenation of the CLS token and the mean of all patch tokens in the output of the ViT). For all results, see Extended Table~\ref{tab:results-full}.
}\label{tab:results}
\centering
\scalebox{0.8}{%
\begin{tabular}{|l|c|lllllllllllll|l|}
\hline
Model name & \#WSIs & pc10  & bach  & brcs  & bkhs  & crc   & glsn  & mhst  & pc    & c16   & pnd   & cnsp  & mnsc  & HEST  & Avg. \\\hline
\textbf{\pt}                         & 92k  & $\textbf{.900}$ & $\textbf{.904}$ & $\textbf{.646}$ &          $.802$ &          $.966$ & $\textbf{.807}$ &          $.828$ & $\textbf{.951}$ &          $.868$ &          $.651$ & $\textbf{.662}$ & $\textbf{.708}$ &          $.415$ & $\textbf{.778}$ \\
UNI-2                                & 350k & $\textbf{.885}$ & $\textbf{.924}$ & $\textbf{.651}$ & $\textbf{.863}$ & $\textbf{.970}$ &          $.777$ &          $.829$ & $\textbf{.951}$ & $\textbf{.873}$ & $\textbf{.666}$ &          $.626$ &          $.644$ & $\textbf{.431}$ & $\textbf{.776}$ \\
\textbf{\nki}                        & 92k  & $\textbf{.882}$ &          $.889$ &          $.615$ &          $.793$ & $\textbf{.967}$ & $\textbf{.823}$ &          $.831$ &          $.948$ & $\textbf{.872}$ &          $.643$ &          $.629$ &          $.656$ & $\textbf{.425}$ & $\textbf{.767}$ \\
Virchow2                             & 3.1M &          $.835$ &          $.890$ &          $.633$ &          $.818$ &          $.966$ &          $.791$ & $\textbf{.865}$ &          $.938$ &          $.860$ &          $.646$ &          $.640$ &          $.674$ &          $.403$ &          $.766$ \\
\textbf{\tcga}                       & 12k  &          $.803$ & $\textbf{.907}$ &          $.639$ &          $.840$ & $\textbf{.967}$ &          $.790$ &          $.815$ &          $.931$ &          $.869$ &          $.656$ &          $.625$ &          $.664$ &          $.412$ &          $.763$ \\
Kaiko-B8                             & 29k  &          $.799$ &          $.876$ &          $.641$ & $\textbf{.842}$ &          $.960$ &          $.761$ &          $.830$ &          $.920$ &          $.836$ &          $.650$ & $\textbf{.644}$ &          $.686$ &          $.391$ &          $.757$ \\
\textbf{tcga-100M}                   & 12k  &          $.789$ &          $.873$ &          $.619$ &          $.814$ & $\textbf{.968}$ & $\textbf{.798}$ &          $.808$ &          $.928$ & $\textbf{.870}$ & $\textbf{.675}$ &          $.622$ &          $.656$ &          $.415$ &          $.757$ \\
H-Optimus-0                          & 500k &          $.831$ &          $.752$ &          $.620$ &          $.813$ &          $.962$ &          $.769$ & $\textbf{.850}$ &          $.943$ &          $.847$ & $\textbf{.672}$ & $\textbf{.644}$ & $\textbf{.687}$ & $\textbf{.425}$ &          $.755$ \\
Prov\_GigaPath                       & 171k &          $.853$ &          $.794$ &          $.626$ & $\textbf{.846}$ &          $.959$ &          $.727$ &          $.831$ &          $.944$ &          $.812$ &          $.657$ &          $.628$ & $\textbf{.688}$ &          $.405$ &          $.752$ \\
Hibou-L                              & 1.1M &          $.825$ &          $.792$ & $\textbf{.643}$ &          $.767$ &          $.954$ &          $.766$ & $\textbf{.850}$ & $\textbf{.949}$ &          $.852$ &          $.654$ & $\textbf{.646}$ &          $.668$ &          $.397$ &          $.751$ \\
UNI                                  & 100k &          $.833$ &          $.797$ &          $.613$ &          $.808$ &          $.954$ &          $.759$ &          $.841$ &          $.937$ &          $.854$ &          $.662$ &          $.627$ &          $.662$ &          $.391$ &          $.749$ \\
vitg14 (nat. img.)                   & 0    &          $.721$ &          $.724$ &          $.578$ &          $.783$ &          $.943$ &          $.740$ & $\textbf{.855}$ &          $.881$ &          $.500$ &          $.509$ &          $.565$ &          $.614$ &          $.351$ &          $.674$ \\
vitg14 (initial)                     & 0    &          $.652$ &          $.474$ &          $.413$ &          $.425$ &          $.754$ &          $.459$ &          $.578$ &          $.763$ &          $.526$ &          $.304$ &          $.462$ &          $.432$ &          $.166$ &          $.493$ \\\hline
\end{tabular}}
\end{table}

The \nki model trained on the TCGA and NKI-80k WSIs (92k WSIs in total) slightly surpasses Virchow2, and is just 0.009 behind UNI-2 (see Table~\ref{tab:results}) despite being trained on 4$\times$ fewer WSIs (92k vs. 350k).
Note that UNI-2~\cite{uni2_huggingface} was released in Jan. 2025 as a successor of UNI~\cite{chen2024uni}.
Despite UNI-2 having used significantly more data for training, our models demonstrate a comparable and sometimes superior performance on the considered downstream tasks.
Unfortunately, we could not add Atlas~\cite{ATD2025} to our evaluation because we did not have the model weights for it.

Finally, our post-trained model \pt demonstrated a superior average accuracy to all other models in the benchmark and surpassed UNI-2 with an average margin of 0.002.
For this evaluation, all the images were resized from their original size specified in Table~\ref{tab:datasets} to \size{392}, instead of resizing them to \size{224} as for all other models.
The results (Table~\ref{tab:results}) indicate that the high-resolution post-training improved the base \nki model especially significantly on the segmentation metrics, CoNSeP and MoNuSAC.
Notably, on the PCam (10 shots) task, which is derived from PatchCamelyon by reducing the training set to just ten random tiles per class (20 in total) for every evaluation run and averaging the test accuracy over 50 training runs, this model achieves the balanced accuracy of $0.90$ and surpasses all other evaluated models.
However, the performance on the Camelyon16 and HEST tasks has significantly degraded, which needs further investigation.
Overall, this still resulted in the absolute best-performing model among all the evaluated models.

To ensure a fair comparison, we additionally evaluated UNI~\cite{chen2024uni} on all downstream tasks with resized images. (In~\cite{chen2024uni}, they mention a fine-tuning performed with larger \size{512} images but without the details about their procedure.) However, the performance of UNI only degraded on images resized to \size{512} (see row `UNI/512' in Extended Table~\ref{tab:results-full}), e.g., 0.89 on PCam, which suggests that their fine-tuning procedure was of a different nature than ours.
We also evaluated UNI-2 in the same way as \pt, on \size{392} images; however, that also yielded slightly lower performance (see row `UNI-2/392' in Extended Table~\ref{tab:results-full}), which again suggests that UNI-2 does not benefit from evaluating on larger images at higher resolution.

Notably, all evaluated pathology FMs surpass the baseline ViT-g14 model trained on natural images (`vitg14 (nat. img.)' in Table~\ref{tab:results}) with a large margin, which highlights the importance of developing domain-specific pathology FMs.
However, all pathology FMs in our benchmark perform relatively poorly on MHIST, where the baseline model trained on natural images is the second-best model.
This suggests that there still remains a potential for improvement, which sets a particular aim for future work.
\subsubsection{Ablation experiments}
To measure the effect of the adaptations we made to the baseline training workflow, we performed an ablation study, where we trained several smaller ViT-B14 models (Table~\ref{tab:ablation}).
These training runs were done for 500k iterations with 4 GPUs, a batch size of 64 per GPU, and accumulating gradients over every three training steps, resulting in an effective total batch size of 768.
The first four experiments evaluated the importance of the HSV filter, the KDE regularizer, and the HED augmentations.
Without replacing the default KoLeo regularizer with KDE and without the HSV filter, our training did not converge, thus, we report `n/a' in the first row of Table~\ref{tab:ablation}.
After adding the HSV filter, the training converged to an average accuracy of 0.704 on eva, which was still far from 0.753, obtained with the final config. The HED color augmentations improved the performance on HEST but did not have a large effect on eva. However, we applied it anyway to help make the FMs more robust to different stainings.
\newcommand{\tick}{}
\DeclareRobustCommand{\yes}{%
  \tikz\fill[scale=0.4, color=black]
  (0,.35) -- (.25,0) -- (1,.7) -- (.25,.15) -- cycle;
}
\begin{table}[t]
\centering
\caption{Performance of the models in the ablation study. Each row corresponds to a single ViT-B14 model trained with the modifications specified by check marks.}\label{tab:ablation}
\scalebox{0.8}{%
\begin{tabular}{|cccc:ccc|cc|}
\hline
TCGA & NKI-80k & CPTAC & GTEx & HSV  & KDE  & HED  & eva   & HEST  \\\hline
\yes &         &       &      &      &      &      & n/a   & n/a   \\
\yes &         &       &      & \yes &      &      & 0.704 & 0.367 \\
\yes &         &       &      & \yes & \yes &      & 0.754 & 0.367 \\
\yes &         &       &      & \yes & \yes & \yes & 0.753 & 0.376 \\\hdashline
\yes & \yes    &       &      & \yes & \yes & \yes & 0.759 & 0.374 \\
\yes &         & \yes  &      & \yes & \yes & \yes & 0.744 & \textbf{0.380} \\
\yes &         &       & \yes & \yes & \yes & \yes & 0.750 & 0.363 \\\hdashline
     & \yes    & \yes  & \yes & \yes & \yes & \yes & 0.742 & 0.368 \\
\yes &         & \yes  & \yes & \yes & \yes & \yes & 0.742 & 0.375 \\
\yes & \yes    &       & \yes & \yes & \yes & \yes & 0.750 & 0.368 \\
\yes & \yes    & \yes  &      & \yes & \yes & \yes & 0.765 & 0.373 \\
\yes & \yes    & \yes  & \yes & \yes & \yes & \yes & \textbf{0.768} & 0.375 \\\hline
\end{tabular}}
\end{table}

Next, we added each of NKI-80k, CPTAC, and GTEx, to evaluate their contribution when added to TCGA. (The tiles were sampled from both data sets with equal probability.)
The results (rows 5, 6, and 7 in Table~\ref{tab:ablation}) show that all three had a rather small impact, with NKI-80k bringing the highest average gain in accuracy.
We also ran data ablations relative to the baseline run on all four data sets: TCGA, GTEx, CPTAC, and NKI-80k (the five last rows of Table~\ref{tab:ablation}).
Here, we excluded each data set at a time and trained a ViT-B14 on the remaining three data sets.
The results show that removing GTEx and CPTAC only marginally affected the FM's final performance, while removing TCGA and NKI-80k resulted in a higher loss.

Last, to check whether we could get an FM with a comparable performance to that of \tcga even with less data, we trained the large ViT-g14 model on just 10\% of the TCGA slides ($\sim$1k WSIs). The resulting performance was far lower, e.g., only 0.9 on PCAM after 500k iterations.
We also trained ViT-g14 on 100M distinct tiles randomly sampled from all the 12k TCGA slides, which resulted in a slightly lower performance than that of \tcga (row `\textbf{tcga-100M}' in Extended Table~\ref{tab:results-full}).
\subsubsection{Image segmentation}
Identifying different cell types can be essential not only for making an accurate diagnosis but also for understanding tumor behavior by analyzing the cellular composition of the micro-environment.
In addition to systematically evaluating the FMs' capability to segment and classify cells in images on the CoNSeP and MoNuSAC tasks (evaluated in eva), here, we selected two images from the CoNSeP data set for a clear visual demonstration and performed the standard semantic segmentation procedure implemented in eva for four models: ViT-g14 (natural images), Lunit, Virchow2, and our models \tcga and \pt.
\begin{figure}[t]
    \centering
    \includegraphics[width=\textwidth]{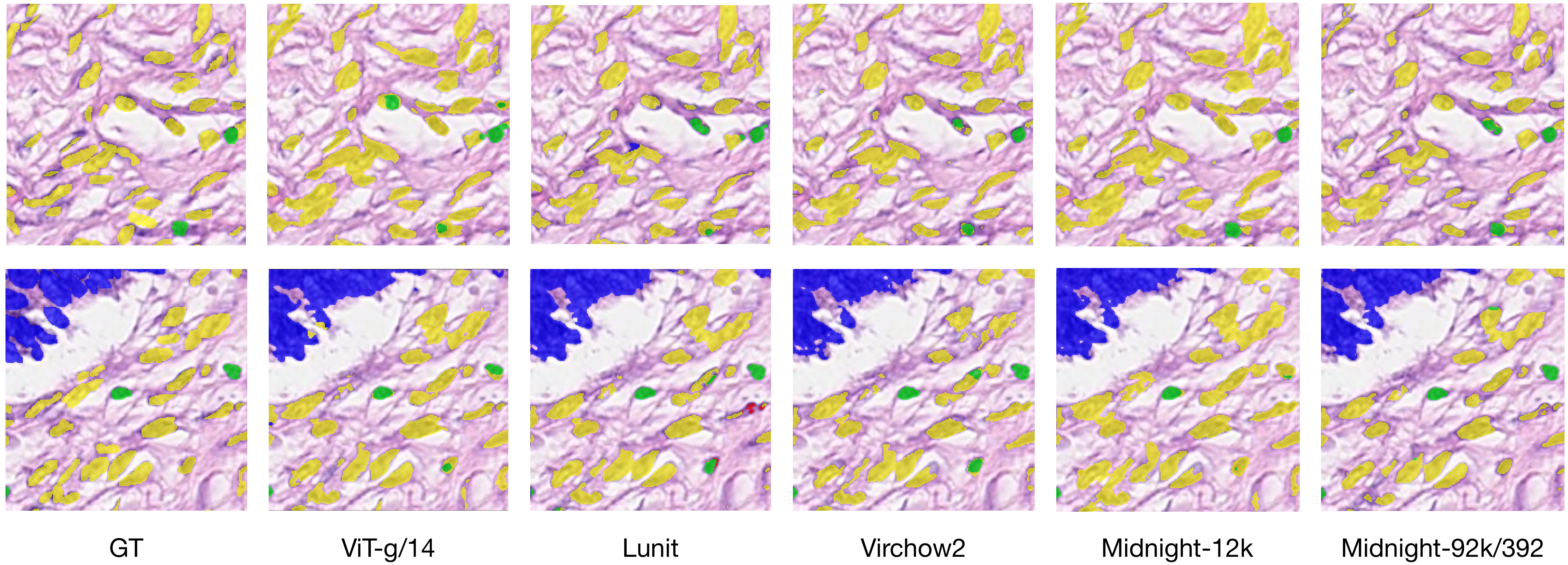}
    \caption{Examples of segmentation performed with different FMs on two tiles from the CoNSeP data set: ViT-g14 (natural images), Lunit, Virchow2, and our models \tcga and \pt. Ground truth is shown on the left side~--- green: inflammatory, blue: epithelial, yellow: spindle-shaped nuclei.}
    \label{fig:cell_segmentation}
\end{figure}
\begin{figure}[!b]
    \centering
    \includegraphics[width=0.85\textwidth]{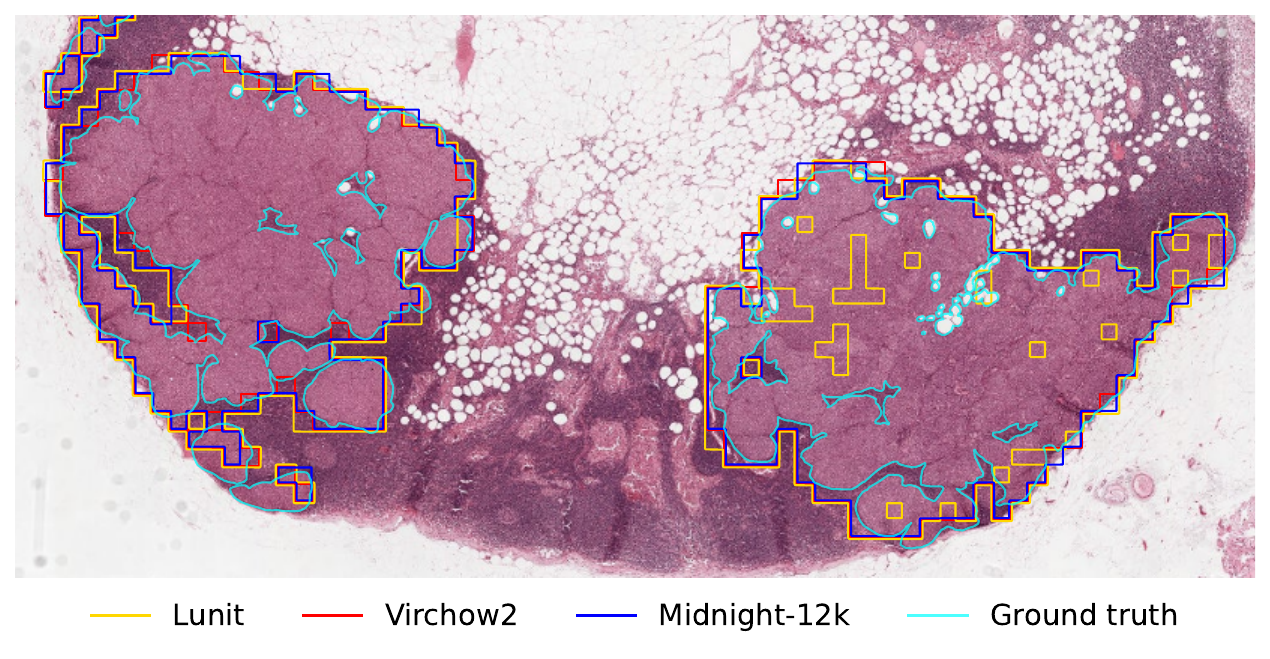}
    \caption{Tile predictions for Lunit, Virchow2, and \tcga and ground truth annotations for slide {\textsf{test\_040}} from Camelyon16.}
    \label{fig:segmentation}
\end{figure}

All pathology FMs produce segmentations (Fig \ref{fig:segmentation}) that are noticeably better than the baseline model, which again highlights the importance of training the FMs on pathology images. Notably, our models produce segmentations that are comparable in quality to those from Virchow2, despite being trained on 34–258×fewer WSIs (12–92k vs. 3.1M).

\subsubsection{Tile-level classification with FM for slide-level segmentation}
Metastasis to regional lymph nodes is an early sign of malignant spread. Thus, detecting lymph node metastasis is crucial in many cancer types, as it upstages the disease and impacts both clinical outcomes and treatment strategies.
To demonstrate how tile-level pathology FMs perform on slide-level tasks, we trained tile-level downstream classifiers to detect lymph node metastases in breast cancer on the Camelyon16 data with three different FMs: Lunit, Virchow2, and \tcga. We applied these classifiers on a randomly picked slide \verb|test_040| (Fig.~\ref{fig:segmentation}).
It can be seen that the predictions for \tcga and Virchow2 are nearly identical and close to the expert annotations. At the same time, the weaker FM Lunit generates far less accurate predictions, which clearly shows the practical importance of using higher-performing FMs to get qualitatively better results on downstream tasks.
\section{Conclusion}
\label{sec:conclusion}
We have presented three new pathology FMs trained on up to two orders of magnitude fewer WSIs than some state-of-the-art models, yet achieving comparable or superior performance on downstream tasks.
We have shown that even with a relatively basic setup, it is possible to train a high-performing pathology FM with far fewer WSIs than one may have previously thought necessary.
We make our \tcga model trained solely on TCGA open for download from~\url{https://huggingface.co/kaiko-ai/midnight} under the MIT license to encourage further research and reproducibility.

With our main results, we do not mean to imply that small data sets are always sufficient to reach state-of-the-art performance. On the contrary, our strong results with magnitudes fewer WSIs show that foundation model training in pathology remains far from saturation. In other words, we believe that there is still significant unrealized potential in today's algorithms~--- potential that can be tapped at truly large scale.
As pathology AI continues to evolve, we believe our work makes a solid contribution to the collective efforts of devising better pathology FMs. This brings us another step closer to achieving real impact in clinics and lowering the burden on pathologists while ultimately improving the quality of provided healthcare.
\section*{Acknowledgements}
We thank the Core Facility for Molecular Pathology and Biobanking of the Netherlands Cancer Institute for providing the NKI-80k digitized slides with approval from the Institutional review board (IRBdm22-188).
We also thank Prof. Dr. Lodewyk Wessels for his support throughout this project.
The GTEx data used for the analyses described in this manuscript were obtained from the GTEx Portal \url{https://gtexportal.org} on October 1, 2024.
The TCGA slides used in this work were generated by the TCGA Research Network: \url{https://www.cancer.gov/tcga}.
We thank Nils Eckstein, Moritz Platscher, Thomas Hufener, and others at kaiko.ai for the helpful discussions and their valuable feedback on this manuscript. We also thank Dang Nguyen, Gianmaria Genetlici, and the rest of the Infrastructure Team at kaiko.ai for maintaining and optimizing the computing resources used in this work.
\bibliographystyle{splncs04}
\bibliography{references.bib}
\newpage
\appendix
\section{Appendix}
\label{sec:appendix}
\subsection{Supplementary Figures}
\setcounter{figure}{0}
\renewcommand{\thefigure}{A\arabic{figure}}
\begin{figure}
    \centering
    \includegraphics[width=0.54\textwidth]{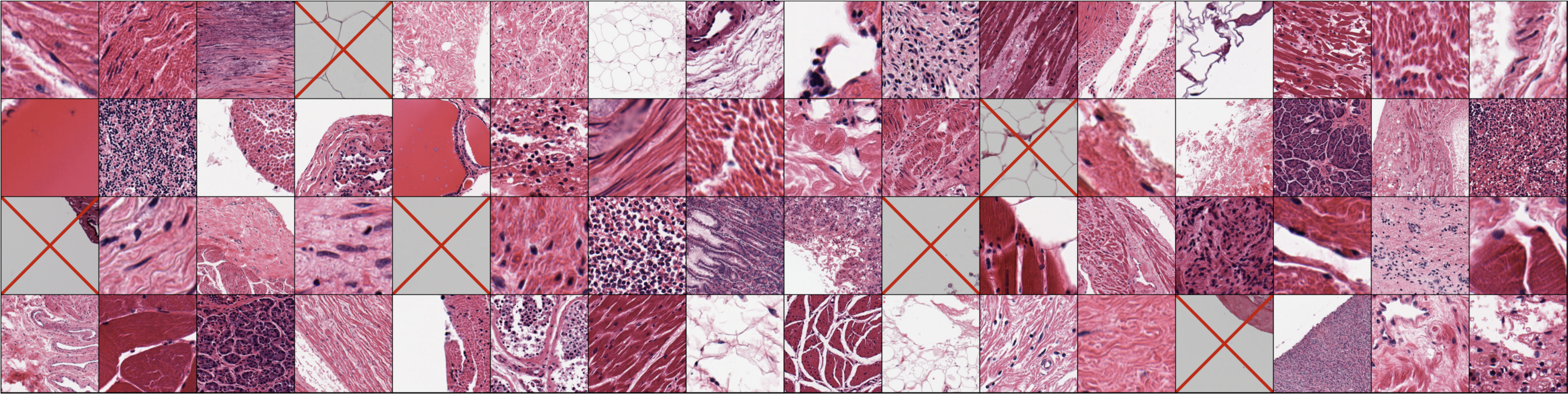}
    \includegraphics[width=0.45\textwidth]{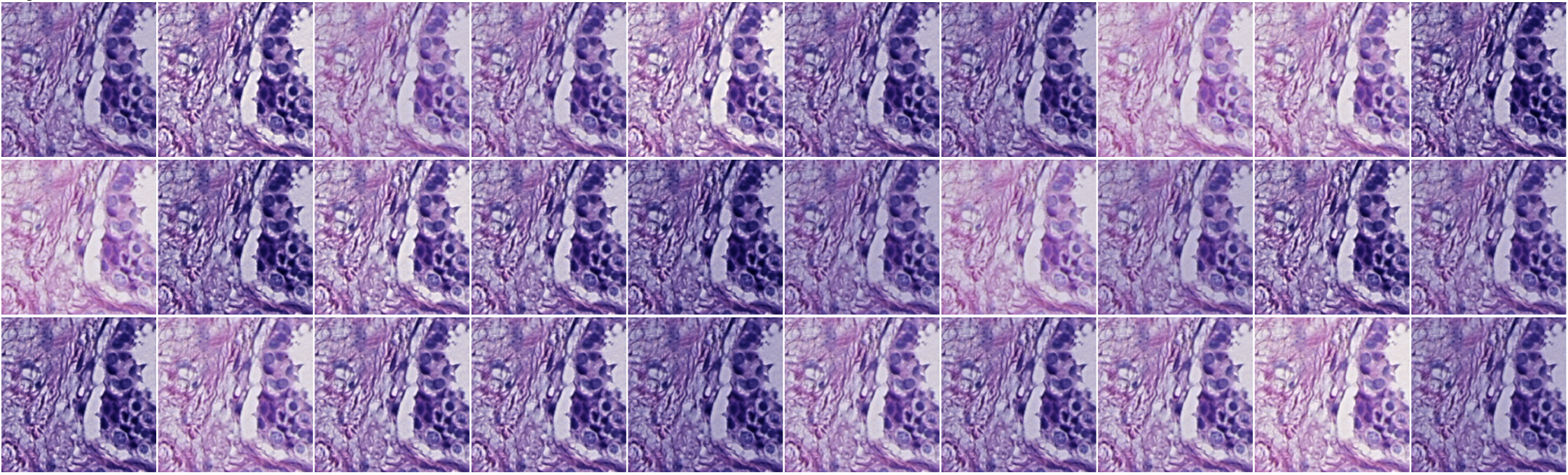}
    \caption{\textbf{Left:} Random tiles cropped from TCGA FFPE slides at 0.5µm/px that passed the HSV filter and those filtered out.
    \textbf{Right:} Random HED augmentations applied to a single tile sampled from the BACH data set.}
    \label{fig:img_preprocessing}
\end{figure}
\subsection{Extended Tables}
\setcounter{table}{0}
\renewcommand{\thetable}{A\arabic{table}}
\begin{sidewaystable}
\caption{Performance of all evaluated FMs on the evaluated downstream tasks and their average.
PCam (10 shots) is a tile-level classification task derived from PCam where the training set is reduced to just ten random tiles per class (20 tiles in total).
We report balanced accuracy for the classification tasks, dice score (no background) for semantic segmentation (CoNSeP and MoNuSAC), and the average Pearson correlation for the nine HEST regression tasks.
All models were evaluated in two variants on the classification tasks, using CLS+Mean token embeddings (default) and CLS only. (For the latter, the respective rows are highlighted in light grey.) The models are sorted by the average performance with the CLS+Mean tokens, and the top three values for each task are highlighted in bold. (The evaluations with the CLS tokens, rows in grey, are not taken into account for this sorting.) Each subscript for the tasks in eva represents the standard deviation of the sample mean computed over multiple runs of the respective downstream task. (*) For HEST, the subscripts represent standard deviations of the predicted gene expressions, averaged across the nine HEST tasks. (**) For the average (last column), the subscripts represent standard deviations averaged across all 13 evaluated downstream tasks.
}\label{tab:results-full}
\rowcolors{3}{white}{gray!20}
\centering
\scriptsize
\renewcommand{\arraystretch}{1.2}
\setlength{\tabcolsep}{3pt}
\scalebox{0.78}{%
\begin{tabular}{|l|c|*{13}{c}|c|}
\hline
Model name & \#WSIs & PCam & BACH & BRACS & BreaKHis & CRC-100K & Gleason & MHIST & PCam & Cam16 & Panda & CoNSeP & MoNuSAC & HEST & Avg. \\
           &         & (10 shots) &         &  &  &  &  &  &  & (small) & (small) &  &  & (*) & (**) \\\hline
\textbf{\pt}                         & 92k  & $\textbf{.900}_{\pm.005}$ &          $.904_{\pm.002}$ & $\textbf{.646}_{\pm.001}$ &          $.802_{\pm.000}$ &          $.966_{\pm.000}$ & $\textbf{.807}_{\pm.001}$ &          $.828_{\pm.000}$ & $\textbf{.951}_{\pm.000}$ &          $.868_{\pm.002}$ &          $.651_{\pm.002}$ & $\textbf{.662}_{\pm.001}$ & $\textbf{.708}_{\pm.002}$ &          $.415_{\pm.038}$ & $\textbf{.778}_{\pm.004}$ \\
\textbf{\pt \texttt{[CLS]}}          & 92k  &          $.900_{\pm.005}$ &          $.906_{\pm.001}$ &          $.642_{\pm.001}$ &          $.850_{\pm.000}$ &          $.964_{\pm.000}$ &          $.809_{\pm.001}$ &          $.825_{\pm.000}$ &          $.951_{\pm.000}$ &          $.831_{\pm.002}$ &          $.633_{\pm.002}$ &          $.663_{\pm.001}$ &          $.707_{\pm.002}$ &          $.384_{\pm.044}$ &          $.774_{\pm.005}$ \\
UNI-2                                & 350k & $\textbf{.885}_{\pm.005}$ & $\textbf{.924}_{\pm.002}$ & $\textbf{.651}_{\pm.004}$ & $\textbf{.863}_{\pm.002}$ & $\textbf{.970}_{\pm.000}$ &          $.777_{\pm.003}$ &          $.829_{\pm.001}$ & $\textbf{.951}_{\pm.000}$ & $\textbf{.873}_{\pm.002}$ & $\textbf{.666}_{\pm.003}$ &          $.626_{\pm.000}$ &          $.644_{\pm.002}$ & $\textbf{.431}_{\pm.039}$ & $\textbf{.776}_{\pm.005}$ \\
UNI-2 \texttt{[CLS]}                 & 350k &          $.887_{\pm.005}$ &          $.914_{\pm.002}$ &          $.661_{\pm.002}$ &          $.860_{\pm.001}$ &          $.965_{\pm.000}$ &          $.778_{\pm.002}$ &          $.823_{\pm.000}$ &          $.949_{\pm.001}$ &          $.868_{\pm.002}$ &          $.659_{\pm.003}$ &          $.628_{\pm.001}$ &          $.644_{\pm.002}$ &          $.414_{\pm.037}$ &          $.773_{\pm.005}$ \\
UNI-2/392                            & 350k &          $.827_{\pm.006}$ & $\textbf{.928}_{\pm.002}$ & $\textbf{.667}_{\pm.003}$ &          $.810_{\pm.001}$ & $\textbf{.967}_{\pm.000}$ &          $.787_{\pm.003}$ & $\textbf{.850}_{\pm.000}$ &          $.929_{\pm.001}$ & $\textbf{.882}_{\pm.004}$ &          $.664_{\pm.003}$ &          $.629_{\pm.001}$ &          $.659_{\pm.001}$ &          $.424_{\pm.041}$ & $\textbf{.771}_{\pm.005}$ \\
UNI-2/392 \texttt{[CLS]}             & 350k &          $.821_{\pm.006}$ &          $.917_{\pm.003}$ &          $.663_{\pm.003}$ &          $.829_{\pm.001}$ &          $.965_{\pm.000}$ &          $.791_{\pm.002}$ &          $.849_{\pm.001}$ &          $.927_{\pm.001}$ &          $.858_{\pm.003}$ &          $.653_{\pm.004}$ &          $.629_{\pm.001}$ &          $.659_{\pm.001}$ &          $.407_{\pm.043}$ &          $.767_{\pm.005}$ \\
\textbf{\nki}                        & 92k  & $\textbf{.882}_{\pm.005}$ &          $.889_{\pm.003}$ &          $.615_{\pm.003}$ &          $.793_{\pm.001}$ & $\textbf{.967}_{\pm.000}$ & $\textbf{.823}_{\pm.001}$ &          $.831_{\pm.000}$ &          $.948_{\pm.000}$ & $\textbf{.872}_{\pm.004}$ &          $.643_{\pm.003}$ &          $.629_{\pm.000}$ &          $.656_{\pm.003}$ & $\textbf{.425}_{\pm.032}$ &          $.767_{\pm.004}$ \\
\textbf{\nki \texttt{[CLS]}}         & 92k  &          $.876_{\pm.006}$ &          $.896_{\pm.002}$ &          $.616_{\pm.004}$ &          $.789_{\pm.001}$ &          $.966_{\pm.000}$ &          $.820_{\pm.002}$ &          $.811_{\pm.000}$ &          $.950_{\pm.000}$ &          $.861_{\pm.005}$ &          $.625_{\pm.003}$ &          $.629_{\pm.000}$ &          $.656_{\pm.002}$ &          $.392_{\pm.039}$ &          $.761_{\pm.005}$ \\
Virchow2                             & 3.1M &          $.835_{\pm.006}$ &          $.890_{\pm.004}$ &          $.633_{\pm.003}$ &          $.818_{\pm.001}$ &          $.966_{\pm.000}$ &          $.791_{\pm.004}$ & $\textbf{.865}_{\pm.001}$ &          $.938_{\pm.001}$ &          $.860_{\pm.001}$ &          $.646_{\pm.003}$ &          $.640_{\pm.001}$ &          $.674_{\pm.002}$ &          $.403_{\pm.046}$ &          $.766_{\pm.006}$ \\
Virchow2 \texttt{[CLS]}              & 3.1M &          $.851_{\pm.007}$ &          $.884_{\pm.004}$ &          $.624_{\pm.003}$ &          $.823_{\pm.004}$ &          $.966_{\pm.000}$ &          $.778_{\pm.006}$ &          $.861_{\pm.000}$ &          $.936_{\pm.001}$ &          $.865_{\pm.001}$ &          $.656_{\pm.004}$ &          $.639_{\pm.001}$ &          $.676_{\pm.002}$ &          $.398_{\pm.040}$ &          $.766_{\pm.006}$ \\
\textbf{\tcga}                       & 12k  &          $.803_{\pm.007}$ & $\textbf{.907}_{\pm.001}$ &          $.639_{\pm.002}$ &          $.840_{\pm.000}$ & $\textbf{.967}_{\pm.000}$ &          $.790_{\pm.001}$ &          $.815_{\pm.000}$ &          $.931_{\pm.000}$ &          $.869_{\pm.003}$ &          $.656_{\pm.003}$ &          $.625_{\pm.000}$ &          $.664_{\pm.001}$ &          $.412_{\pm.036}$ &          $.763_{\pm.004}$ \\
\textbf{\tcga \texttt{[CLS]}}        & 12k  &          $.791_{\pm.007}$ &          $.904_{\pm.001}$ &          $.644_{\pm.001}$ &          $.841_{\pm.000}$ &          $.966_{\pm.000}$ &          $.801_{\pm.001}$ &          $.807_{\pm.000}$ &          $.930_{\pm.000}$ &          $.850_{\pm.004}$ &          $.663_{\pm.002}$ &          $.626_{\pm.001}$ &          $.663_{\pm.001}$ &          $.395_{\pm.040}$ &          $.760_{\pm.005}$ \\
Kaiko-B8                             & 29k  &          $.799_{\pm.007}$ &          $.876_{\pm.004}$ &          $.641_{\pm.004}$ & $\textbf{.842}_{\pm.003}$ &          $.960_{\pm.000}$ &          $.761_{\pm.006}$ &          $.830_{\pm.000}$ &          $.920_{\pm.000}$ &          $.836_{\pm.001}$ &          $.650_{\pm.003}$ & $\textbf{.644}_{\pm.000}$ &          $.686_{\pm.001}$ &          $.391_{\pm.048}$ &          $.757_{\pm.007}$ \\
Kaiko-B8 \texttt{[CLS]}              & 29k  &          $.786_{\pm.008}$ &          $.872_{\pm.006}$ &          $.617_{\pm.005}$ &          $.825_{\pm.004}$ &          $.957_{\pm.000}$ &          $.748_{\pm.004}$ &          $.828_{\pm.001}$ &          $.917_{\pm.001}$ &          $.831_{\pm.006}$ &          $.642_{\pm.003}$ &          $.643_{\pm.001}$ &          $.686_{\pm.001}$ &          $.373_{\pm.052}$ &          $.748_{\pm.007}$ \\
\textbf{tcga-100M}                   & 12k  &          $.789_{\pm.007}$ &          $.873_{\pm.002}$ &          $.619_{\pm.002}$ &          $.814_{\pm.001}$ & $\textbf{.968}_{\pm.000}$ & $\textbf{.798}_{\pm.001}$ &          $.808_{\pm.000}$ &          $.928_{\pm.000}$ &          $.870_{\pm.004}$ & $\textbf{.675}_{\pm.003}$ &          $.622_{\pm.000}$ &          $.656_{\pm.002}$ &          $.415_{\pm.038}$ &          $.757_{\pm.005}$ \\
\textbf{tcga-100M \texttt{[CLS]}}    & 12k  &          $.774_{\pm.007}$ &          $.864_{\pm.001}$ &          $.615_{\pm.001}$ &          $.779_{\pm.000}$ &          $.967_{\pm.000}$ &          $.799_{\pm.000}$ &          $.792_{\pm.000}$ &          $.927_{\pm.000}$ &          $.852_{\pm.007}$ &          $.667_{\pm.005}$ &          $.622_{\pm.001}$ &          $.656_{\pm.001}$ &          $.396_{\pm.041}$ &          $.747_{\pm.005}$ \\
H-Optimus-0                          & 500k &          $.831_{\pm.007}$ &          $.752_{\pm.004}$ &          $.620_{\pm.006}$ &          $.813_{\pm.003}$ &          $.962_{\pm.000}$ &          $.769_{\pm.004}$ & $\textbf{.850}_{\pm.001}$ &          $.943_{\pm.001}$ &          $.847_{\pm.003}$ & $\textbf{.672}_{\pm.003}$ & $\textbf{.644}_{\pm.002}$ & $\textbf{.687}_{\pm.001}$ & $\textbf{.425}_{\pm.037}$ &          $.755_{\pm.006}$ \\
H-Optimus-0 \texttt{[CLS]}           & 500k &          $.824_{\pm.007}$ &          $.757_{\pm.003}$ &          $.615_{\pm.004}$ &          $.808_{\pm.004}$ &          $.956_{\pm.002}$ &          $.771_{\pm.003}$ &          $.842_{\pm.002}$ &          $.942_{\pm.001}$ &          $.838_{\pm.004}$ &          $.670_{\pm.004}$ &          $.644_{\pm.001}$ &          $.685_{\pm.002}$ &          $.415_{\pm.038}$ &          $.751_{\pm.006}$ \\
Prov\_GigaPath                       & 171k &          $.853_{\pm.006}$ &          $.794_{\pm.002}$ &          $.626_{\pm.002}$ & $\textbf{.846}_{\pm.001}$ &          $.959_{\pm.000}$ &          $.727_{\pm.004}$ &          $.831_{\pm.000}$ &          $.944_{\pm.000}$ &          $.812_{\pm.004}$ &          $.657_{\pm.003}$ &          $.628_{\pm.001}$ & $\textbf{.688}_{\pm.001}$ &          $.405_{\pm.045}$ &          $.752_{\pm.005}$ \\
Prov\_GigaPath \texttt{[CLS]}        & 171k &          $.852_{\pm.006}$ &          $.766_{\pm.004}$ &          $.616_{\pm.002}$ &          $.821_{\pm.002}$ &          $.951_{\pm.000}$ &          $.720_{\pm.005}$ &          $.831_{\pm.001}$ &          $.942_{\pm.001}$ &          $.791_{\pm.006}$ &          $.660_{\pm.003}$ &          $.626_{\pm.002}$ &          $.687_{\pm.002}$ &          $.393_{\pm.047}$ &          $.743_{\pm.006}$ \\
Hibou-L                              & 1.1M &          $.825_{\pm.006}$ &          $.792_{\pm.004}$ &          $.643_{\pm.005}$ &          $.767_{\pm.002}$ &          $.954_{\pm.001}$ &          $.766_{\pm.001}$ & $\textbf{.850}_{\pm.001}$ & $\textbf{.949}_{\pm.001}$ &          $.852_{\pm.003}$ &          $.654_{\pm.002}$ & $\textbf{.646}_{\pm.001}$ &          $.668_{\pm.001}$ &          $.397_{\pm.052}$ &          $.751_{\pm.006}$ \\
Hibou-L \texttt{[CLS]}               & 1.1M &          $.804_{\pm.007}$ &          $.811_{\pm.006}$ &          $.637_{\pm.007}$ &          $.740_{\pm.001}$ &          $.933_{\pm.002}$ &          $.763_{\pm.001}$ &          $.839_{\pm.000}$ &          $.952_{\pm.001}$ &          $.823_{\pm.003}$ &          $.634_{\pm.005}$ &          $.645_{\pm.001}$ &          $.668_{\pm.001}$ &          $.388_{\pm.048}$ &          $.740_{\pm.007}$ \\
UNI                                  & 100k &          $.833_{\pm.006}$ &          $.797_{\pm.006}$ &          $.613_{\pm.003}$ &          $.808_{\pm.004}$ &          $.954_{\pm.001}$ &          $.759_{\pm.007}$ &          $.841_{\pm.001}$ &          $.937_{\pm.002}$ &          $.854_{\pm.003}$ &          $.662_{\pm.004}$ &          $.627_{\pm.001}$ &          $.662_{\pm.003}$ &          $.391_{\pm.049}$ &          $.749_{\pm.007}$ \\
UNI \texttt{[CLS]}                   & 100k &          $.815_{\pm.007}$ &          $.791_{\pm.009}$ &          $.593_{\pm.001}$ &          $.789_{\pm.005}$ &          $.948_{\pm.002}$ &          $.757_{\pm.005}$ &          $.840_{\pm.001}$ &          $.938_{\pm.001}$ &          $.822_{\pm.006}$ &          $.655_{\pm.004}$ &          $.627_{\pm.000}$ &          $.659_{\pm.004}$ &          $.386_{\pm.051}$ &          $.740_{\pm.007}$ \\
UNI/512                              & 100k &          $.755_{\pm.008}$ &          $.891_{\pm.004}$ &          $.624_{\pm.004}$ &          $.751_{\pm.004}$ &          $.951_{\pm.001}$ &          $.757_{\pm.007}$ &          $.831_{\pm.000}$ &          $.890_{\pm.001}$ &          $.811_{\pm.005}$ &          $.647_{\pm.002}$ &          $.620_{\pm.001}$ &          $.671_{\pm.003}$ &          $.380_{\pm.045}$ &          $.737_{\pm.007}$ \\
UNI/512 \texttt{[CLS]}               & 100k &          $.737_{\pm.008}$ &          $.877_{\pm.008}$ &          $.612_{\pm.001}$ &          $.732_{\pm.004}$ &          $.950_{\pm.000}$ &          $.754_{\pm.010}$ &          $.814_{\pm.002}$ &          $.883_{\pm.001}$ &          $.814_{\pm.006}$ &          $.654_{\pm.002}$ &          $.621_{\pm.001}$ &          $.658_{\pm.003}$ &          $.364_{\pm.046}$ &          $.728_{\pm.007}$ \\
Phikon                               & 12k  &          $.826_{\pm.007}$ &          $.744_{\pm.003}$ &          $.579_{\pm.002}$ &          $.715_{\pm.006}$ &          $.946_{\pm.001}$ &          $.743_{\pm.005}$ &          $.824_{\pm.000}$ &          $.919_{\pm.002}$ &          $.822_{\pm.006}$ &          $.648_{\pm.003}$ &          $.624_{\pm.001}$ &          $.644_{\pm.002}$ &          $.377_{\pm.048}$ &          $.724_{\pm.007}$ \\
Phikon \texttt{[CLS]}                & 12k  &          $.820_{\pm.007}$ &          $.735_{\pm.004}$ &          $.568_{\pm.015}$ &          $.713_{\pm.004}$ &          $.942_{\pm.002}$ &          $.729_{\pm.003}$ &          $.804_{\pm.001}$ &          $.923_{\pm.000}$ &          $.809_{\pm.008}$ &          $.644_{\pm.004}$ &          $.623_{\pm.001}$ &          $.644_{\pm.003}$ &          $.367_{\pm.050}$ &          $.717_{\pm.008}$ \\
Phikon-v2                            & 60k  &          $.756_{\pm.007}$ &          $.737_{\pm.004}$ &          $.607_{\pm.003}$ &          $.725_{\pm.002}$ &          $.953_{\pm.000}$ &          $.753_{\pm.001}$ &          $.796_{\pm.000}$ &          $.900_{\pm.000}$ &          $.807_{\pm.003}$ &          $.634_{\pm.003}$ &          $.626_{\pm.000}$ &          $.645_{\pm.003}$ &          $.391_{\pm.045}$ &          $.718_{\pm.006}$ \\
Phikon-v2 \texttt{[CLS]}             & 60k  &          $.741_{\pm.008}$ &          $.734_{\pm.003}$ &          $.600_{\pm.003}$ &          $.716_{\pm.003}$ &          $.939_{\pm.001}$ &          $.755_{\pm.002}$ &          $.784_{\pm.000}$ &          $.893_{\pm.000}$ &          $.803_{\pm.003}$ &          $.631_{\pm.002}$ &          $.626_{\pm.001}$ &          $.645_{\pm.003}$ &          $.375_{\pm.041}$ &          $.711_{\pm.005}$ \\
Lunit                                & 36k  &          $.763_{\pm.007}$ &          $.785_{\pm.006}$ &          $.627_{\pm.002}$ &          $.759_{\pm.005}$ &          $.943_{\pm.001}$ &          $.758_{\pm.004}$ &          $.785_{\pm.000}$ &          $.905_{\pm.001}$ &          $.759_{\pm.011}$ &          $.604_{\pm.003}$ &          $.600_{\pm.000}$ &          $.630_{\pm.001}$ &          $.362_{\pm.063}$ &          $.714_{\pm.008}$ \\
Lunit \texttt{[CLS]}                 & 36k  &          $.753_{\pm.008}$ &          $.782_{\pm.008}$ &          $.614_{\pm.004}$ &          $.750_{\pm.007}$ &          $.938_{\pm.001}$ &          $.747_{\pm.003}$ &          $.779_{\pm.001}$ &          $.901_{\pm.001}$ &          $.730_{\pm.006}$ &          $.610_{\pm.004}$ &          $.599_{\pm.001}$ &          $.629_{\pm.001}$ &          $.353_{\pm.054}$ &          $.707_{\pm.008}$ \\
vitg14 (nat. img.)                   & 0    &          $.721_{\pm.006}$ &          $.724_{\pm.002}$ &          $.578_{\pm.003}$ &          $.783_{\pm.001}$ &          $.943_{\pm.000}$ &          $.740_{\pm.002}$ & $\textbf{.855}_{\pm.001}$ &          $.881_{\pm.001}$ &          $.500_{\pm.010}$ &          $.509_{\pm.006}$ &          $.565_{\pm.001}$ &          $.614_{\pm.001}$ &          $.351_{\pm.042}$ &          $.674_{\pm.006}$ \\
vitg14 (nat. img.) \texttt{[CLS]}    & 0    &          $.719_{\pm.005}$ &          $.725_{\pm.003}$ &          $.583_{\pm.002}$ &          $.832_{\pm.001}$ &          $.935_{\pm.000}$ &          $.744_{\pm.002}$ &          $.862_{\pm.001}$ &          $.874_{\pm.001}$ &          $.507_{\pm.004}$ &          $.382_{\pm.016}$ &          $.564_{\pm.001}$ &          $.614_{\pm.001}$ &          $.342_{\pm.043}$ &          $.668_{\pm.006}$ \\
vitg14 (initial)                     & 0    &          $.652_{\pm.006}$ &          $.474_{\pm.005}$ &          $.413_{\pm.007}$ &          $.425_{\pm.003}$ &          $.754_{\pm.003}$ &          $.459_{\pm.002}$ &          $.578_{\pm.004}$ &          $.763_{\pm.001}$ &          $.526_{\pm.008}$ &          $.304_{\pm.003}$ &          $.462_{\pm.001}$ &          $.432_{\pm.005}$ &          $.166_{\pm.087}$ &          $.493_{\pm.010}$ \\
vitg14 (initial) \texttt{[CLS]}      & 0    &          $.649_{\pm.006}$ &          $.473_{\pm.002}$ &          $.411_{\pm.004}$ &          $.427_{\pm.008}$ &          $.748_{\pm.003}$ &          $.464_{\pm.003}$ &          $.569_{\pm.000}$ &          $.755_{\pm.002}$ &          $.566_{\pm.006}$ &          $.308_{\pm.002}$ &          $.461_{\pm.003}$ &          $.428_{\pm.004}$ &          $.172_{\pm.089}$ &          $.495_{\pm.010}$ \\\hline
\end{tabular}}
\end{sidewaystable}
\end{document}